\documentclass[letterpaper, 10 pt, conference]{ieeeconf}  % Comment this line out if you need a4paper

\IEEEoverridecommandlockouts                              % This command is only needed if 
\usepackage{graphics} % for pdf, bitmapped graphics files
\usepackage{amsmath}
\usepackage{amssymb}
\usepackage{mathtools}
\usepackage{bm}
\usepackage[T2A,T1]{fontenc}
\usepackage{hyperref}
\usepackage{microtype}
\usepackage{graphicx}
\usepackage{booktabs}
\usepackage{eso-pic} % used by \AddToShipoutPicture
\usepackage{url}
\usepackage{times} % assumes new font selection scheme installed
\usepackage{algorithm} % For algorithms
\usepackage{algpseudocode}
\usepackage{balance}
\usepackage{amssymb}
\usepackage{xspace}

\def\eg{{\it e.g.}}

\def\ie{{\it i.e.}}

\title{\LARGE \bf
Path Planning using Instruction-Guided Probabilistic Roadmaps
}

\author{Jiaqi Bao$^{1}$ and Ryo Yonetani$^{1}$% <-this % stops a space
\thanks{$^{1}$ JB and RY are with CyberAgent, Inc., Tokyo, Japan.
        {\tt\small jiaqibao1230@gmail.com, yonetani\_ryo@cyberagent.co.jp}}%
}

\begin{document}

\maketitle

\thispagestyle{empty}
\pagestyle{empty}

%%%%%%%%%%%%%%%%%%%%%%%%%%%%%%%%%%%%%%%%%%%%%%%%%%%%%%%%%%%%%%%%%%%%%%%%%%%%%%%%
\begin{abstract}
This work presents a novel data-driven path planning algorithm named Instruction-Guided Probabilistic Roadmap (IG-PRM). Despite the recent development and widespread use of mobile robot navigation, the safe and effective travels of mobile robots still require significant engineering effort to take into account the constraints of robots and their tasks. With IG-PRM, we aim to address this problem by allowing robot operators to specify such constraints through natural language instructions, such as ``aim for wider paths'' or ``mind small gaps''. The key idea is to convert such instructions into embedding vectors using large-language models (LLMs) and use the vectors as a condition to predict instruction-guided cost maps from occupancy maps. By constructing a roadmap based on the predicted costs, we can find instruction-guided paths via the standard shortest path search. Experimental results demonstrate the effectiveness of our approach on both synthetic and real-world indoor navigation environments.
\end{abstract}

\section{Introduction}
Autonomous mobile robots have begun to be used not only in conventional factories and warehouses but also in everyday environments such as restaurants and retail stores. Whether carrying items or sweeping the floor, having a mobile robot effectively and safely accomplish a task requires in-depth knowledge of both the robot and the task. Consider a catering robot working in a restaurant. To avoid spilling food and drinks, even small steps must be avoided. How about a robot in a retail store? It might be better to prioritize narrow aisles so as not to disturb shoppers, or go through wide streets instead so that more people can see the ads printed on the robot. But how much ``small'' steps or ``narrow'' aisles would be acceptable? Answering this question requires a sufficient understanding of the robot's specifications.

\begin{figure}[t]
    \centering
    \includegraphics[width=.9\linewidth]{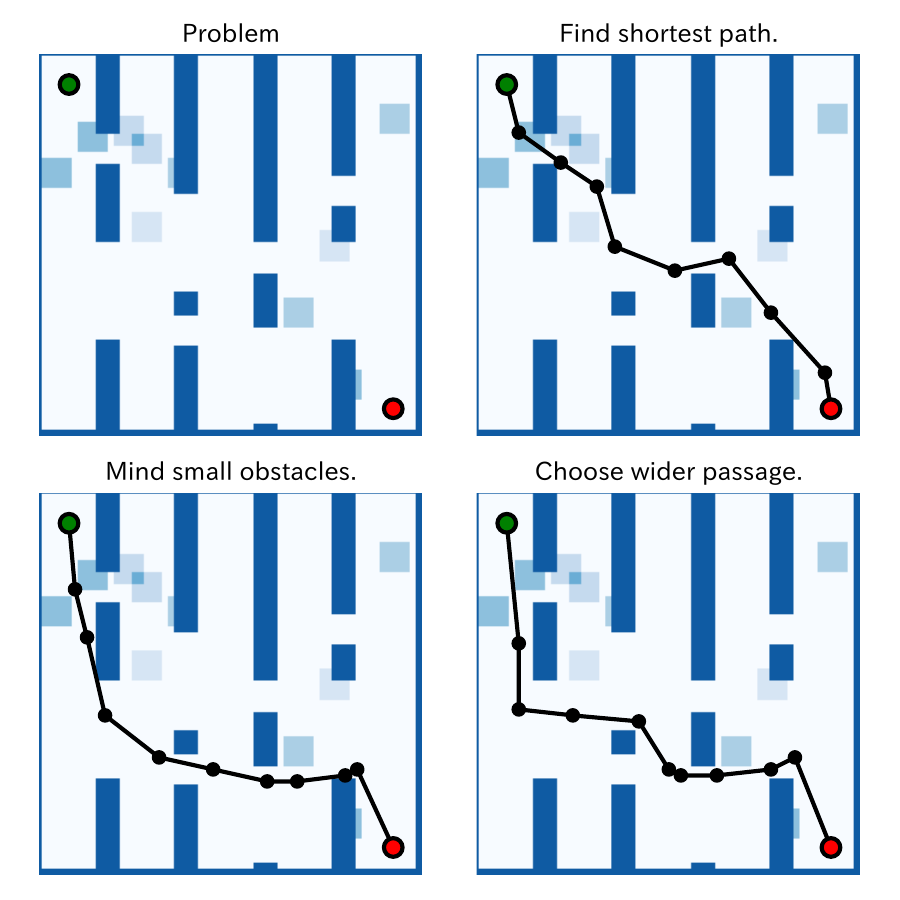}
    \caption{\textbf{Instruction-Guided Path Planning.} Given natural language instructions such as ``choose wider passage,'' our planner finds a collision-free path from the start to the goal while taking into account the instructions.}
    \label{fig:teaser}
\end{figure}

On the one hand, we envision a future mobile robot capable of understanding ambiguous task instructions given by humans in natural language, such as ``mind small gaps for your safety''. This will make it easier for non-experts to install and operate mobile robot applications. On the other hand, we also aim to make the best use of existing, well-established navigation frameworks consisting of global planners and other functionalities as a separate module, rather than replacing the complete framework into a single, black-box machine learning model converting instructions into paths. We believe such modularity is critical to keep the whole system maintainable and easy to debug.

\begin{figure*}[t]
    \centering
    \includegraphics[width=\linewidth]{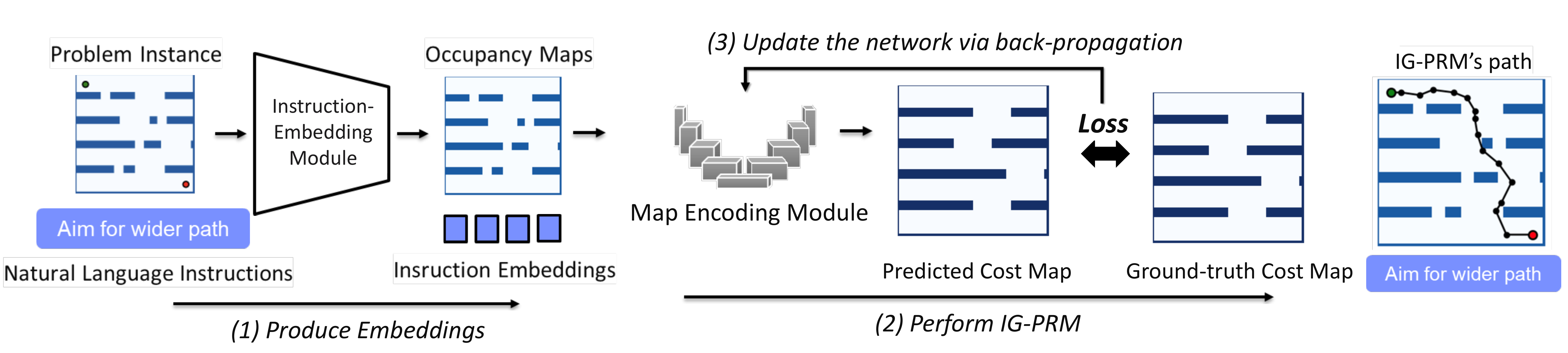}
    \caption{\textbf{Schematic Diagram of IG-PRM.} (1) Convert natural language instructions into embeddings. (2) Combine occupancy maps with embeddings to predict instruction-guided cost maps, where the cost prediction network is trained in a supervised fashion. (3) Use the predicted cost maps for producing instruction-guided paths.}
    \label{fig:framework}
\end{figure*}

In this work, we develop a novel data-driven path planning algorithm named \textbf{Instruction-Guided Probabilistic Roadmaps (IG-PRM)}. As shown in Fig~\ref{fig:teaser}, IG-PRM takes occupancy maps and text instructions as inputs and plans a feasible path that satisfies the instructions. The key idea is to learn a neural network that encodes occupancy maps and instructions into an instruction-guided cost map (see also Fig~\ref{fig:framework}.) The cost map is then used to adaptively construct a roadmap, on which instruction-guided paths can be found via a standard shortest path search algorithm. As such, IG-PRM can act as a drop-in replacement for global planners which only modulates cost maps to allow for text instructions.

We systematically evaluate the effectiveness of IG-PRM with both synthetic and real-world indoor navigation environments~\cite{xia2018gibson,dobrevski2020adaptive}. Experimental results demonstrate our planner's superiority in path optimality and alignment with natural language instructions compared to popular path planners such as probabilistic roadmaps (PRM)~\cite{kavraki1996probabilistic,karaman2011sampling}.

\section{Related work}
\paragraph{Data-Driven Path Planning}
Path planning (in particular point-to-point shortest path planning) refers to the task of finding a low-cost, collision-free path from start to goal points in a map, which has a long history in robotics and AI~\cite{hart1968formal,abd2015comprehensive,gonzalez2015review}. Sampling-based path planning, such as Probabilistic Roadmaps~\cite{kavraki1996probabilistic}, Rapidly-exploring Random Trees (RRT)~\cite{lavalle1998rapidly}, or their asymptotically optimal versions~\cite{karaman2011sampling}, is a popular approach to the path planning tasks especially when the robotic state space is either continuous or high-dimensional. This approach approximates the state space by a fixed number of samples (\ie, nodes of roadmap graphs or random trees), where biased sampling around potential solution paths can improve planning efficiency. Machine learning has been a promising tool for learning how to bias the sampling from expert demonstration data~\cite{ichter2018learning,chen2019learning,ichter2020learned}. Nevertheless, no prior work on such data-driven path planning can incorporate natural language instructions into their planning results.

\paragraph{LLMs for Robotic Applications}
Large language models (LLMs) have been used for various applications including robotics~\cite{zeng2023large}. Existing work utilizes LLMs for task planning and action selection~\cite{ahn2022can,joublin2024copal,hu2024deploying,liu2023llmp}, replanning and self recovery~\cite{shirasaka2023selfrecovery,skreta2024replan}, or human-mediated navigation~\cite{xiao2023llm,dai2024think}. Another interesting work is ROS-GPT~\cite{koubaa2023rosgpt} which integrates OpenAI's ChatGPT with the Robot Operating System (ROS) 2, facilitating the generation of command prompts from natural language inputs. In contrast, our work is the first to utilize LLMs for modulating cost maps in the context of data-driven path planning.

\section{Preliminaries}
\subsection{Path Planning in Continuous State Spaces}
\label{subsec:path planning}
Let $X=[0, 1] \subseteq \mathbb{R}^d$ denote the bounded continuous state space, where we consider $d=2$ for the robot's 2-D positions as the state. The state space is partitioned into free space $X_{\text{free}}$, where the robot can move without encountering obstacles, and obstacle regions $X_{\text{obs}}= X\setminus X_{\text{free}}$ where movement is impeded. The start and goal positions of the robot are given by $x_{\text{start}}, x_{\text{goal}} \in X_{\text{free}}$ respectively. A path $\sigma: [0,1] \rightarrow X$ is a continuous mapping from the unit interval to the state space, with the properties that $\sigma(0) = x_{\text{start}}$ and $\sigma(1) = x_{\text{goal}}$. The path is said to be feasible if $\forall \tau \in [0,1], \sigma(\tau) \in X_{\text{free}}$.

The objective of the path planning problem is to find a feasible path $\sigma^*$ from $x_{\text{start}}$ to $x_{\text{goal}}$ with its cost minimum among all possible paths from the start to the goal. The cost typically reflects user-defined criteria such as path length, energy expenditure, or adherence to safety margin.

\subsection{Probabilistic Roadmap}
Probabilistic Roadmap (PRM)~\cite{kavraki1996probabilistic,karaman2011sampling} is a popular approach to path planning in the continuous state space. It approximates the free space $X_{\text{free}}$ by a graph $G=(V, E),\; V\subset X_{\text{free}},\; E\subset V\times V$ called roadmap, and finds the shortest path on the graph via standard path search algorithms (\eg, the Dijkstra algorithm). PRM has the guarantee for probabilistic completeness and asymptotic optimality; the more nodes we sample from the free space, the higher chance we obtain to find feasible and optimal paths if they exist.

\subsection{Problem Setup}
In this work, we consider a variant of the path planning problem in Sec.~\ref{subsec:path planning}. Our problem instance consists of the state space $X, X_{\text{free}}, X_{\text{obs}}$, start and goal $x_{\text{start}}, x_{\text{goal}}$, and an instruction given in natural languages by robotic operators. Here we introduce the following assumptions: (1) Instructions are short sentences like ``aim for wider paths'' or ``find the shortest route'', explicitly or implicitly indicate some constraints that the robot must comply with; (2) Domain experts can provide a ground-truth cost map function $C: X\rightarrow \mathbb{R}_{\geq 0}$ that maps each state into non-negative cost value imposed by the constraints, but (3) \emph{this cost mapping is demonstrated only for a limited number of problem instances}. For a new problem instance with unknown cost mapping, the solution is a discretized path on a roadmap, \ie, $P=(v_1,\dots,v_T)\;$, $v_1=x_{\text{start}}$ and $v_T=x_{\text{goal}}$, $\forall t \in [1,\dots, T-1],\;(v_t, v_{t+1})\in E$, with its total hidden cost $\sum_{v\in P} C(v)$ being minimum among other possible paths from $x_{\text{start}}$ to $x_{\text{goal}}$.
 
\section{Path Planning using IG-PRM}
As illustrated in Fig.~\ref{fig:framework}, our proposed approach is straightforward; learning a cost prediction model from a collection of problem instances and natural language instructions with the known cost mapping. As implied in its name, IG-PRM utilizes the predicted cost maps for path planning by PRMs to derive a solution path aligned with the instruction that induced the cost.

\subsection{Instruction-Guided Cost Maps}
Our cost prediction network consists of two modules: the instruction embedding module and the map encoding module. First, the instruction embedding module encodes instruction sentences into fixed-size vector embedding (\ie, instruction embedding). We utilize an off-the-shelf LLM that has been trained on a large-scale text corpus, designed to measure semantic similarities between sentences.

After applying dimensionality reduction, the instruction embedding is concatenated along the channel axis with occupancy maps that represent the obstacle regions in $X_{\text{obs}}$, and fed to the map encoding module to produce cost maps of the same size. Any neural network architecture can be employed as long as it facilitates image-to-image translations, such as U-Net~\cite{ronneberger2015u} or other fully convolutional semantic segmentation networks~\cite{garcia2017review}. It is important to note that our cost prediction network does not use start and goal positions map to estimate cost maps, unlike some prior works~\cite{ichter2018learning,yonetani2021path,takahashi2019learning,bhardwaj2017learning}. This feature allows our planner to generate cost maps just once for each combination of instructions and environments, naturally supporting multi-query settings (\ie, solving multiple path planning problems using the same roadmap graph) just like the PRM framework.

While the instruction embedding module can be pre-trained and kept frozen, the map encoding module needs to be trained from scratch. We believe that the diversity of instructions is the key to learning a robust model. To this end, we leverage off-the-shelf LLMs such as ChatGPT~\cite{openai2023gpt4} to paraphrase the same instruction in multiple ways for data augmentation, such as ``Choose the narrower routes for your journey'', ``Navigate through the more confined spaces'' and ``Opt for the less wide pathways'' for the task that needs to choose narrower passages. Given a collection of occupancy maps, instruction embeddings, and ground-truth cost maps derived by experts, we employ the binary cross-entropy loss to measure the discrepancy between the predicted and ground-truth costs and train the map encoding module via standard back-propagation.

\subsection{Instruction-Guided Roadmaps}
Given a new problem instance consisting of occupancy maps and instructions, the learned map encoding module predicts a cost map, which is then used in both node sampling and edge expansion procedures of PRM as follows to construct an instruction-guided roadmap.
\begin{itemize}
 \item \textbf{Node sampling.} Unlike the standard PRM that adopts uniform sampling, the proposed IG-PRM samples nodes from the free space with the probability inversely proportional to the predicted costs.
 \item \textbf{Edge expansion.} While we connect nodes with their neighbors in a standard way based on whether there are no obstacles between them, we assign edge costs proportional to both the edge length and predicted costs. Specifically, we first sample line segments connecting the two nodes at regular intervals and then accumulate the predicted cost values at each sampled point to provide the edge cost. 
\end{itemize}

Once the roadmap is given, we perform a standard Dijkstra search from the start to the goal to find a solution path with a minimum total cost, which should also be aware of the original instruction.

\section{Experiments}
We evaluate the performance of the proposed IG-PRM and other path planning techniques in synthetic and real-world navigation environments.

\begin{figure*}[t]
    \centering
    \includegraphics[width=\linewidth]{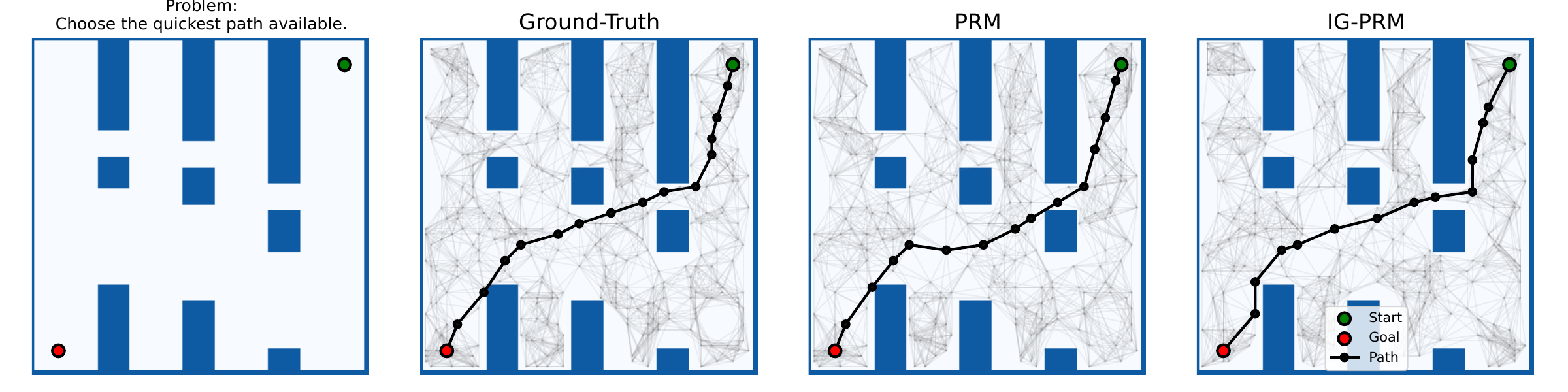}    
    \includegraphics[width=\linewidth]{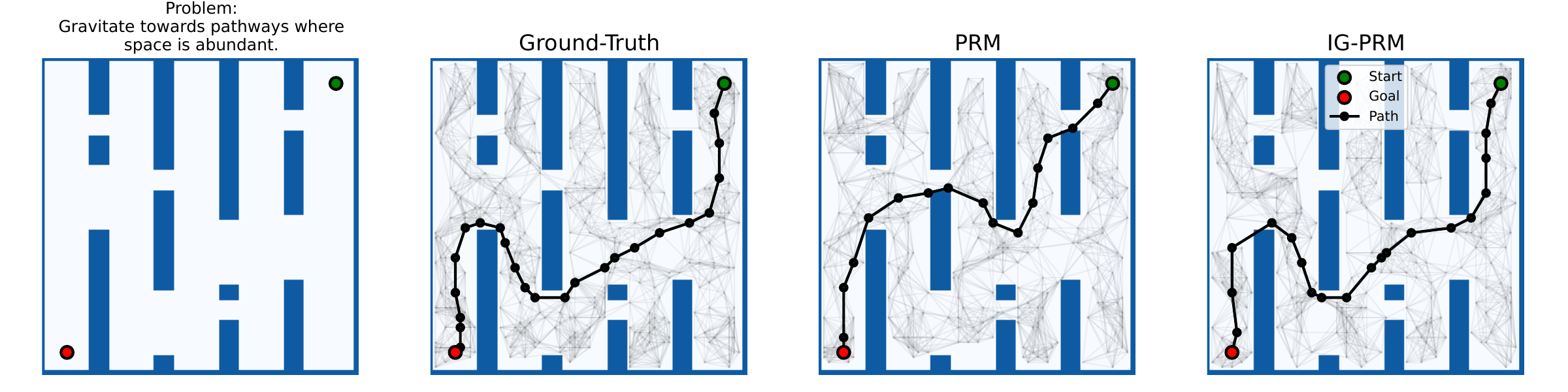}   
    \includegraphics[width=\linewidth]{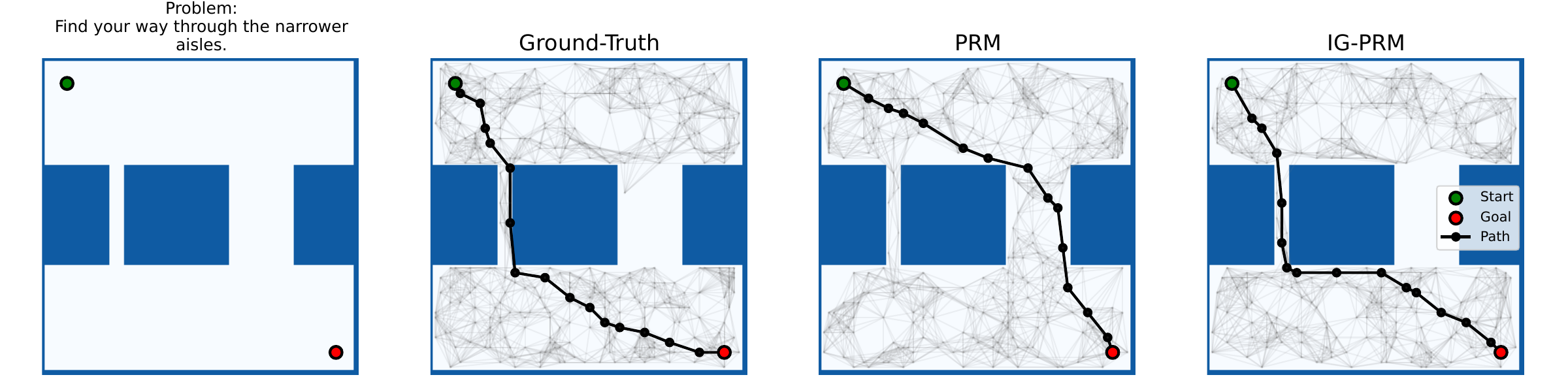} 
    \caption{\textbf{Qualitative Results on Synthetic Instances.} Roadmap and solution paths are visualized with gray and black lines, respectively. Obstacle regions are colored in blue. The start and goal points are marked with green and red circles, respectively.}
    \label{fig:qualitative_results}
\end{figure*}

\begin{figure*}[t]
    \centering
    \includegraphics[width=\linewidth]{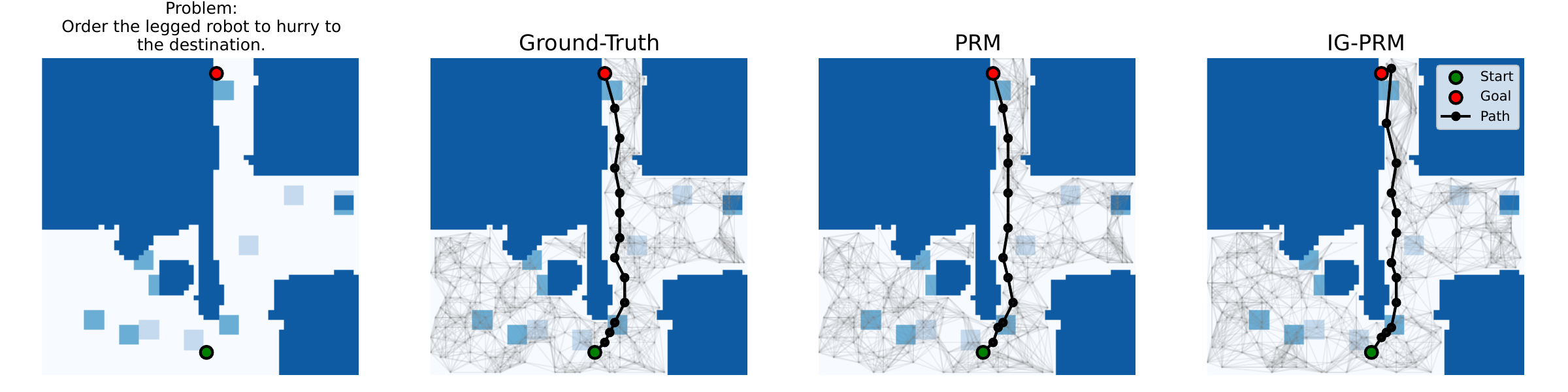} 
    \includegraphics[width=\linewidth]{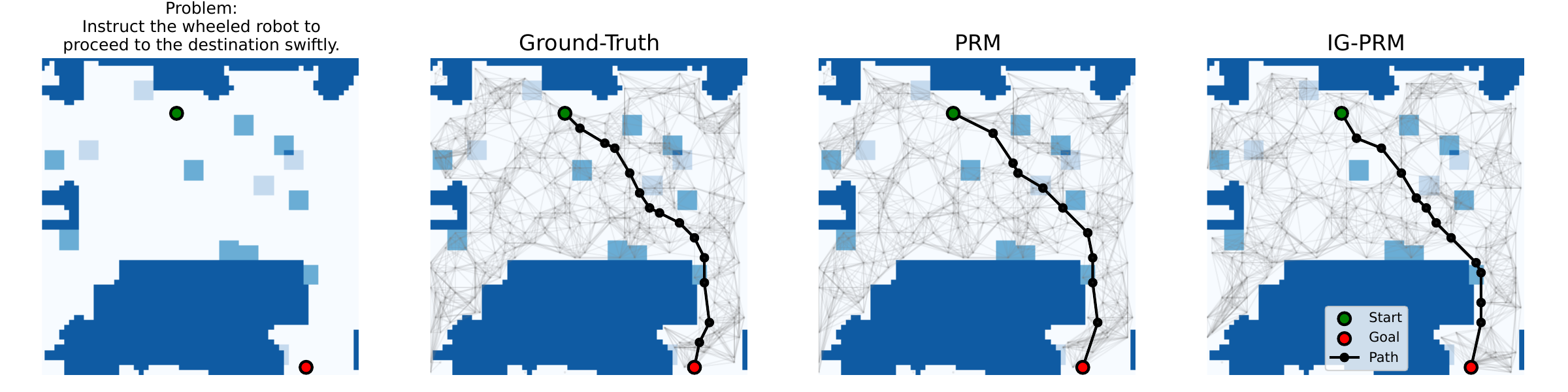} 
    \includegraphics[width=\linewidth]{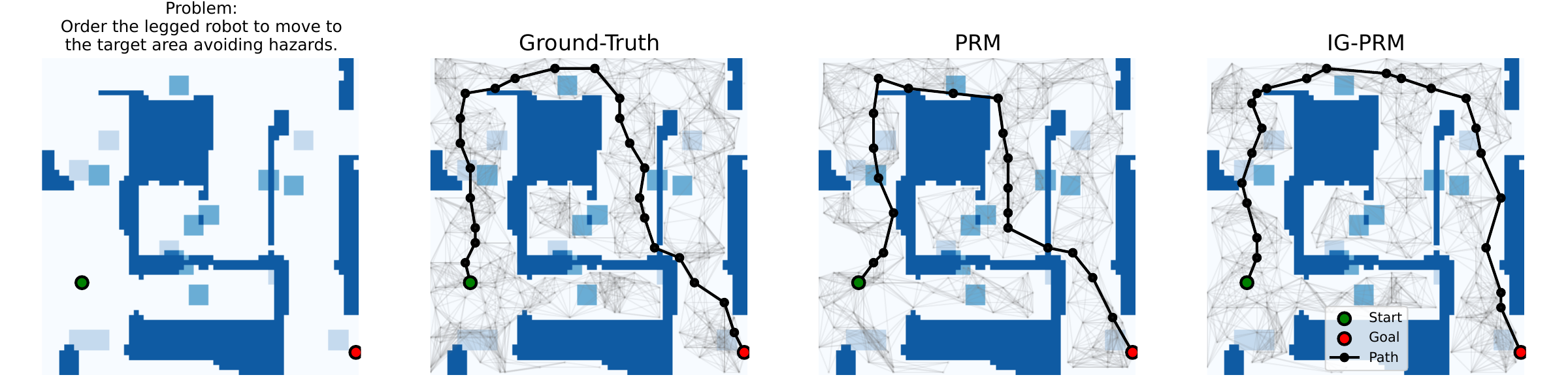} 
    \includegraphics[width=\linewidth]{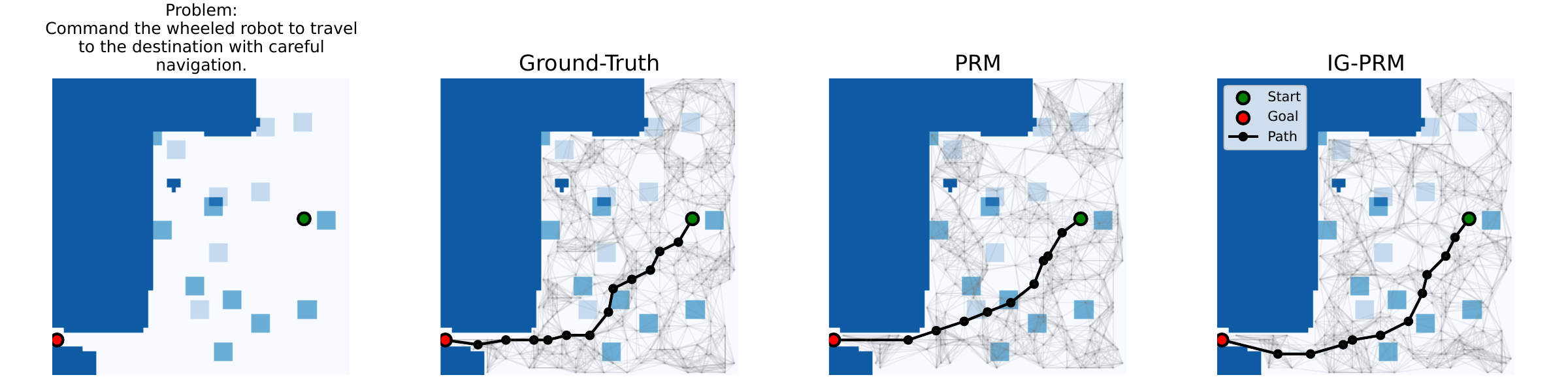} 
    \caption{\textbf{Qualitative Results on IN2D Instances.} Roadmap and solution paths are visualized with gray and black lines, respectively. Obstacle regions are colored in blue. The start and goal points are marked with green and red circles, respectively.}
    \label{fig:qualitative_results_in2d}
\end{figure*}

\subsection{Environments}
\paragraph{Synthetic Environment}
We synthesized a sufficiently controlled environment with a random distribution of walls with variable-width passages in a 64x64-pixel environmental map. As shown in Fig.~\ref{fig:qualitative_results}, each problem instance contains one to four walls of horizontal or vertical directions, where each wall has narrower and wider passages. Start and goal positions are determined randomly so that they are opposite each other at the four corners. We generated 132 diverse instruction sentences that express either one of the following navigation preferences in various ways: a) prioritize narrower passages, b) prioritize wider passages, and c) seek the shortest paths. With these instructions, we assume a practical scenario where robots in crowded places need to avoid people or get more opportunities to interact with them, or anyway hurry to their goal to complete their task.

\paragraph{Indoor Navigation Environment}
Moreover, we extended the Indoor Navigation 2D Dataset (IN2D)~\cite{dobrevskiskocaj2020in2d} as a real-world environment. This dataset is derived from Gibson Database~\cite{xiazamirhe2018gibsonenv}, containing 17 maps of indoor office/residential spaces with scans of real-world walls and furniture as obstacles (see also Fig.~\ref{fig:qualitative_results_in2d}). We resized each environmental map such that its shorter edge is 128 pixels, and further cropped random 64x64 pixel patches from it to create diverse random problem instances. We further placed 20 random 4x4 square `step'-obstacles each with either one of two different heights (higher or lower), simulating small steps that the robot may or may not be able to easily go over. We created 90 diverse instruction sentences that move a) either wheeled or legged robots to b) reach the destination carefully or rapidly (\eg, \emph{``Instruct the wheeled robot to move to the destination cautiously.''}) We assume that, if required to be careful, legged robots need to avoid higher step-obstacles while wheeled robots must avoid both higher and lower ones, based on their kinematic characteristics.

\paragraph{Ground-truth Cost Maps and Paths} 
For each problem instance, we derived the ground-truth cost map based on a hand-engineered function provided by experts. For synthetic problem instances, we gave a high cost to obstacle regions and assigned the same high cost to the narrower passages if the instructions prioritize wider ones, and did vise versa when we needed to prioritize narrower passages instead. As for the IN2D instances, if the instructions ask robots to move carefully, we gave a high cost to all step-obstacles for wheel-robot cases and only higher step-obstacles for legged-robot cases. As such, expert rules on cost mapping tend to be complicated, and our work aims to learn a cost prediction model from diverse training data instead. Ground-truth paths were obtained using the PRM with 400 nodes on the ground-truth cost maps, which properly parsed and took into account the original intention of the given instructions.

\paragraph{Data Splits} For each environment, we created 800 training, 100 validation, and 200 test problem instances. Half of the test instances shared the same set of instruction sentences with training/validation instances, and the remaining test instances were with new sentences. This way, we aim to evaluate if IG-PRM can generalize to unknown instructions.

\subsection{Implementation Details}
We synthesized diverse instruction sentences using GPT-3.5 and GPT-4o, and converted them into embedding vectors using text-embedding-ada-002 and text-embedding-3-small, all available on Microsoft Azure OpenAI Services.\footnote{\url{https://azure.microsoft.com/products/ai-services/openai-service}} The resulting instruction embeddings are 1536-D vectors, which we further converted to be 128-D using Gaussian random projection. For the map encoding module, we used a U-Net~\cite{ronneberger2015u} with a VGG-16 backbone. The output layer of this network uses a sigmoid activation function to normalize the predicted cost maps. We trained the map encoder using the Adam optimizer for 30 epochs, with the batch size set to 64 and the learning rate to 0.001. During training, we saved the checkpoint that achieved the lowest validation loss for later evaluation. All the implementations were done in Python.

\subsection{Evaluation Metrics}
We evaluated the success rate weighted by path length (SPL)~\cite{anderson2018evaluation} and dynamic time warping distances (DTW)~\cite{muller2007dynamic} between produced and ground-truth paths. While SPL scores measure the validity and cost-efficiency of produced paths in a single unified metric, DTW scores directly measure the similarity of the produced paths against the ground-truth solutions. The planning was regarded as successful if the resulting paths did not overlap with high-cost regions in the ground-truth cost maps. We compared these scores while changing the number of nodes within 50, 150, and 300. 

\begin{figure}[t]
    \centering
    \includegraphics[width=\linewidth]{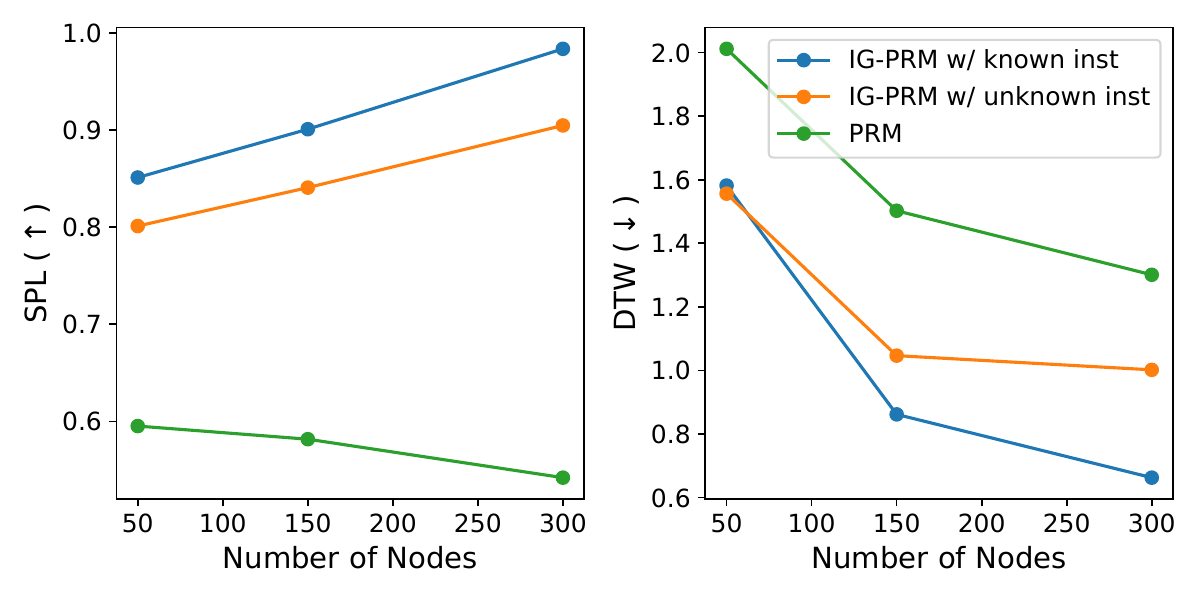}
    \caption{\textbf{Quantitative Results on Synthetic Instances.} We evaluated success rates weighted by path lengths (SPL; the higher, the better) as well as dynamic-time-warping distances (DTW; the lower, the better) against ground-truth paths for each method while varying the number of nodes.}
    \label{fig:plot_passage2_nodes} 
\end{figure}

\begin{figure}[t]
    \centering
    \includegraphics[width=\linewidth]{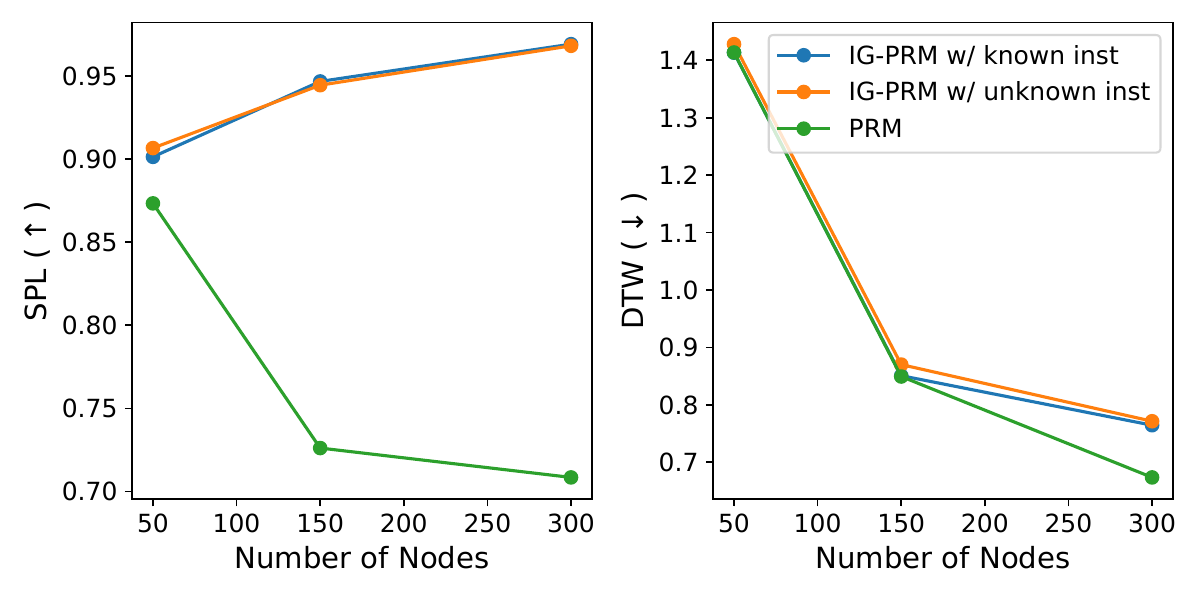}
    \caption{\textbf{Quantitative Results on IN2D Instances.} We evaluated success rates weighted by path lengths (SPL; the higher, the better) as well as dynamic-time-warping distances (DTW; the lower, the better) against ground-truth paths for each method while varying the number of nodes.}
    \label{fig:plot_in2d_nodes}
\end{figure}

\subsection{Results}
\paragraph{Quantitative Evaluation}
We compared the proposed IG-PRM with a standard PRM. Fig.~\ref{fig:plot_passage2_nodes} summarizes SPL and DTW scores against variable numbers of nodes on the synthetic environment instances. As the number of nodes increases, the results show that IG-PRM consistently achieves higher SPL and lower DTW values on both known and unknown instructions. The same trend can be confirmed for IN2D instances in Fig.~\ref{fig:plot_in2d_nodes}, with smaller discrepancies between known and unknown instructions. This result indicates the potential usefulness of practical mobile robotic applications in the real world. On the other hand, PRM overall demonstrated degraded performances (except for the DTW scores with 300 nodes in IN2D instances in Fig.~\ref{fig:plot_in2d_nodes}). This is due to the lack of the ability to take into account instructions, as shortest paths are not necessarily the solution depending on the instructions.

\paragraph{Qualitative Evaluation}
Fig.~\ref{fig:qualitative_results} and Fig.~\ref{fig:qualitative_results_in2d} showcase visual comparisons between IG-PRM and PRM across different scenarios with a fixed number of nodes of 150. As shown in the second and third examples in Fig.~\ref{fig:qualitative_results}, IG-PRM can produce paths while successfully choosing wider/narrower passages based on given instructions, as few nodes and edges are included around the other types of passages in the constructed roadmap. On the IN2D instances, IG-PRM can effectively avoid lower or higher step-obstacles depending on the types of robots (legged for the third and wheeled for the fourth examples in Fig.~\ref{fig:qualitative_results_in2d}) and carefulness requirements mentioned in instructions.

\begin{figure}[t]
    \centering
    \includegraphics[width=\linewidth]{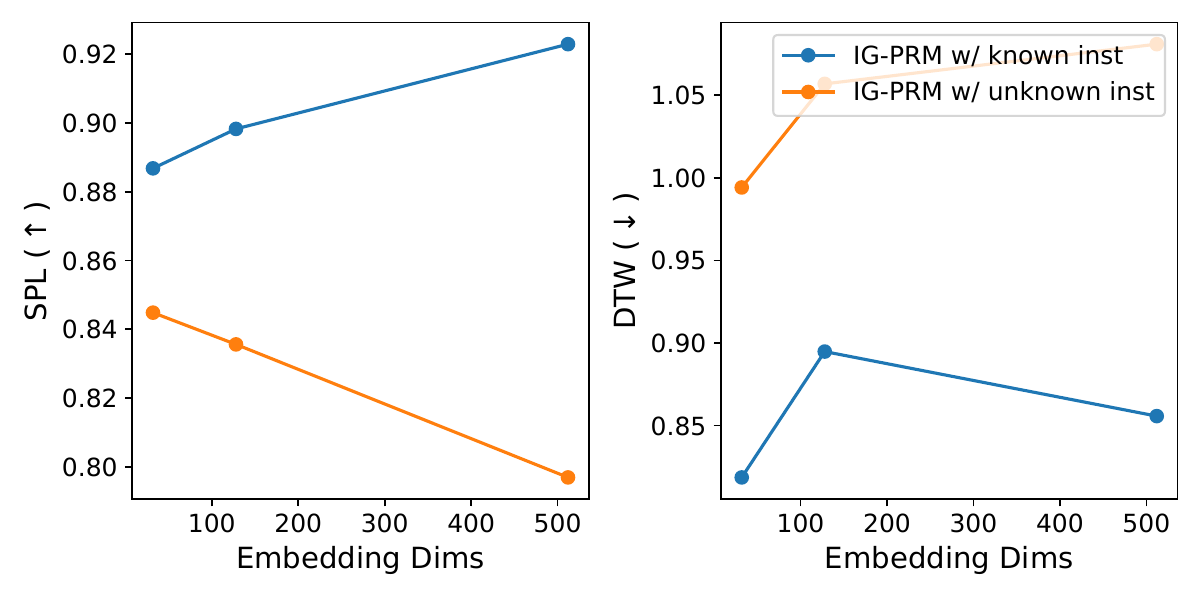}
    \caption{\textbf{Ablation Study.} We evaluated success rates weighted by path lengths (SPL; the higher, the better) as well as dynamic-time-warping distances (DTW; the lower, the better) against ground-truth paths for the proposed method while changing the dimension of embedding vectors for language instructions.}
    \label{fig:plot_passage2_embs}
\end{figure}

\paragraph{Ablation Study}
One unique tunable parameter for IG-PRM is the dimension of instruction embeddings after dimensionality reduction, which affects the generalization ability with respect to instructions.
Fig.~\ref{fig:plot_passage2_embs} reports the ablation study results on known and unknown instruction datasets over varying embedding dimensions. As the dimension increases, the performance on known instruction sets improves significantly. Conversely, for unknown sets, there is a noticeable decline in performance after a certain point, indicating the overfitting issue. 

\paragraph{Computation Time Evaluation}
While we leverage neural networks for planning, IG-PRM can run reasonably fast on CPUs. We measured the computation runtime for running cost prediction and path planning with Intel\textregistered\xspace{} Core\texttrademark\xspace{} i7-1360P. Taking an average of 100 instances, we confirmed it takes 0.15 seconds for the cost prediction (\ie, the time overhead in IG-PRM over the standard PRM) and then 0.23 seconds for the path planning with 300 nodes. The use of GPUs and more optimized implementations can further accelerate the entire pipeline, which we leave for future work.

\subsection{Limitations and Possible Extensions}
Our work has a few limitations. While we learn the cost prediction network with diverse instruction sentences, IG-PRM cannot determine whether a given instruction is relevant to the robot and the task to be addressed, and will `hallucinate' some cost map even if we input a totally unrelated sentence. Moreover, the proposed method cannot properly handle multiple instructions that contradict each other. Toward a safer and more reliable framework, one of the necessary future directions is to have an additional language model to validate input instructions and to ask the user for additional clarifications or interactions if necessary, just like today's LLM-based chatbots often do. Another limitation is that the proposed approach is currently confirmed effective only in 2D environments. Extending it to 3D or higher state spaces would also be a promising direction, where advanced map encoders such as the ones implemented in \cite{ichter2020learned} will be critical. Finally, although PRM was adopted in this work, the cost-map encoder learned by the proposed method can be combined with other standard planners such as A* and RRT.

\section{Conclusion}
We have presented a novel data-driven path planning algorithm named IG-PRM, which can incorporate natural language instructions into the planning framework. We convert the instructions into vector embeddings and use them as additional input to predict instruction-guided costs from occupancy maps. The predicted costs can then be used to adaptively construct roadmaps and find an instruction-guided path via a standard lowest-cost path search. Thanks to this simplicity, IG-PRM can act as a drop-in replacement for global planners such as PRM, allowing us to keep the high modularity of robotic navigation frameworks. Experimental results demonstrated the effectiveness of IG-PRM on both synthetic and real-world navigation environments.

\clearpage

\balance
\bibliographystyle{IEEEtran}
\bibliography{ref.bib}

\end{document}